\documentclass{article}

\usepackage{arxiv}

\usepackage[utf8]{inputenc} % allow utf-8 input
\usepackage[T1]{fontenc}    % use 8-bit T1 fonts
\usepackage{hyperref}       % hyperlinks
\usepackage{url}            % simple URL typesetting
\usepackage{booktabs}       % professional-quality tables
\usepackage{amsfonts}       % blackboard math symbols
\usepackage{nicefrac}       % compact symbols for 1/2, etc.
\usepackage{microtype}      % microtypography
\usepackage{lipsum}
\usepackage{graphicx}
\usepackage{listings}

\title{Tree of Knowledge: an Online Platform for
Learning the Behaviour of Complex Systems}

\author{
  Benedikt T.~Kleppmann\thanks{Use footnote for providing further
    information about author (webpage, alternative
    address)---\emph{not} for acknowledging funding agencies.} \\
    Innovationszentrum Aalen\\
    Anton-Huber-Straße 20\\
    D-73430 Aalen\\

  \texttt{www.treeofknowledge.ai} \\
}

\begin{document}
\maketitle

\begin{abstract}
Many social sciences such as psychology and economics try to learn the behaviour of
complex agents such as humans, organisations and countries. The current statistical methods used for learning this behaviour try to infer generally valid behaviour, but can only
learn from one type of study at a time. Furthermore, only data from carefully designed
studies can be used, as the phenomenon of interest has to be isolated and confounding
factors accounted for. These restrictions limit the robustness and accuracy of insights
that can be gained from social/economic systems. Here we present the online platform
TreeOfKnowledge which implements a new methodology specifically designed for learning complex behaviours from complex systems: agent-based behaviour learning. With
agent-based behaviour learning it is possible to gain more accurate and robust insights as
it does not have the restriction of conventional statistics. It learns agent behaviour from
many heterogenous datasets and can learn from these datasets even if the phenomenon
of interest is not directly observed, but appears deep within complex systems. This new
methodology shows how the internet and advances in computational power allow for more
accurate and powerful mathematical models.
\end{abstract}

% keywords can be removed

\section{Introduction}
Social and economic systems often involve multiple interacting agents (such as people, organisations or countries). Such systems can display very complex behaviours such as emergence, evolution, historicity, out-of-equilibrium dynamics and indeterminacy \cite{Mitleton-Kelly:2003, Arthur:2018,Tesfatsion:2002}. 

Macroscopic models that view the system as a whole cannot capture such behaviour. Instead of considering the behaviour of individual agents, macroscopic models try to find functions that describe the overall behaviour of a uniform mass of agents. For this simplification, strong assumptions have to be made which also have been criticized \cite{Arthur:2005, Heckbert+Baynes+Reeson:2010,Graebner:2015}.

These considerations have led to the increasing popularity of agent-based modelling in fields such as economics \cite{Colander+Holt+Rosse:2004, Deissenberg+Van_der_Hoog+Dawid:2008}, sociology \cite{Macy+Willer:2002}, psychology \cite{Smith+Conrey:2007}, anthropology \cite{Premo:2006}, geography \cite{Heppenstall+Crooks+See+Batty:2011}, ecology \cite{Grimm+Railsback:2005}, biology \cite{Coakley+Smallwood+Holcombe:2006}, medicine \cite{Fullstone+Wood+Holcombe+Battaglia:2015}, and manufacturing \cite{Shen+Wang+Hao:2006}. Agent-based modelling is the approach of simulating a system on the microscopic level. It is able to capture all the complex behaviour and interactions of a system by simulating the evolution of the individual agents over time. How the agents behave and interact is manually defined by a set of rules or by computer code.
The problem therefore lies in knowing the correct behaviour rules. In general, the correct behaviour rules are not known and modellers establish these rules mostly through common sense and guesswork  \cite{Farmer+Foley:2009, Janssen+Ostrom:2006}. Agent-based models are therefore rarely exact enough to be used for making quantitative predictions. These simplified representations of reality are instead used for investigating what microscopic mechanisms give rise to certain macroscopic patterns \cite{Srbljinovic+Skunca:2003, Carley:1996}. 

Investigating what basic mechanisms can create macroscopic phenomena is valuable, but we are interested in accurately modelling these complex economic/social systems. Is there any way of improving our knowledge of the rules that govern the behaviour of humans and other agents?
Fields like psychology have been learning about human behaviour for centuries. The statistical methods they use are however not well adapted for learning the behaviour of these complex agents. The goal is to learn generally valid behaviour, but the statistical methods can only analyse a single dataset from a single study. And meta-analyses can only combine insights from very similar studies \cite{Wolf:1986, Schulze:2004}. Furthermore, much care in the study design has to be taken to isolate a single phenomenon and avoid confounding factors \cite{Breakwell+Smith+Wright:2012, Senn:2012}. 

In this paper we present agent-based behaviour learning, a new methodology specialized on learning the behaviour of complex agents in complex systems. It uses agent-based modelling to learn and validate agent behaviours in any number of situations and with any number of complex datasets. 
This paper describes how agent-based behaviour learning was implemented in the online data analysis platform TreeOfKnowledge. We show how this platform combines the insights from many models and many users to make an increasingly accurate and complete model of social and economic systems. In the final discussion, we compare agent-based behaviour learning to the existing methods in statistics and machine learning and show how it can gain significantly more robust and more accurate insights. Further technical details are presented in the appendix.

%% -- Agent-based Behaviour Learning-------------------------------------------
\section{Agent-based behaviour learning} \label{sec:abbl}
\subsection{Verifying agent behaviour} \label{ssec:verification}

Suppose we already knew the correct behaviour rules for agents, then we could model any scenario with these agents and they would behave correctly.

We can verify if an agent is behaving correctly in a specific scenario with the following consideration: Every dataset contains real-world observations from a specific setting/scenario. An accurate agent-based model of this scenario should reproduce the same values as were observed in the real-word. I.e., if we managed to make an accurate model of such a scenario, we can say that the agents’ behaviours are correct in this scenario.

To verify an agents’ behaviour in many scenarios, we need many datasets. For each of these datasets we model the scenario it was captured in using the same agents. We know that the agents behave correctly in all of these scenarios if all of the models reproduce the same values as their dataset i.e. as was observed in the real-world. 

\subsection{Tree of Knowledge} \label{ssec:tree_of_knowledge}

We will illustrate how agent-based behaviour learning works in practice with the example of the online platform TreeOfKnowledge (\url{www.treeofknowledge.ai}). 

Users can upload datasets to TreeOfKnowledge and model the scenario in which the dataset was captured. All users build their models using the same agents, which come from a central repository. Should an agent not yet exist in this central repository, users can add it. For technical details about the agents, see Appendix~\ref{app:agent_types}.

Users can also modify and extend the behaviour of agents and have these changes automatically fitted to and tested on all models from all users. Like this the agent behaviour is accurately learned from many different models and many different heterogenous datasets simultaneously.

\subsection{Model building} \label{ssec:model_building}

Let’s illustrate the model building with a use case example:

The human agents in TreeOfKnowledge‘s central repository already have several rules on happiness, but there is no rule that considers the influence of someone’s hunger level on their happiness. We however suspect, that hunger could play a role. To test this, we create a new rule which says that “if someone’s hunger-level exceeds 4, then their happiness is reduced by X”. This rule uses the attributes hunger and happiness, which previous users had defined as continuous values between 0 and 10, and X which we define to be between -10 and 10. We would like to learn the unknown parameter X and find out if the new rule improves model accuracy. 

\begin{figure}[t!]
\centering
\includegraphics[width=1.0\textwidth]{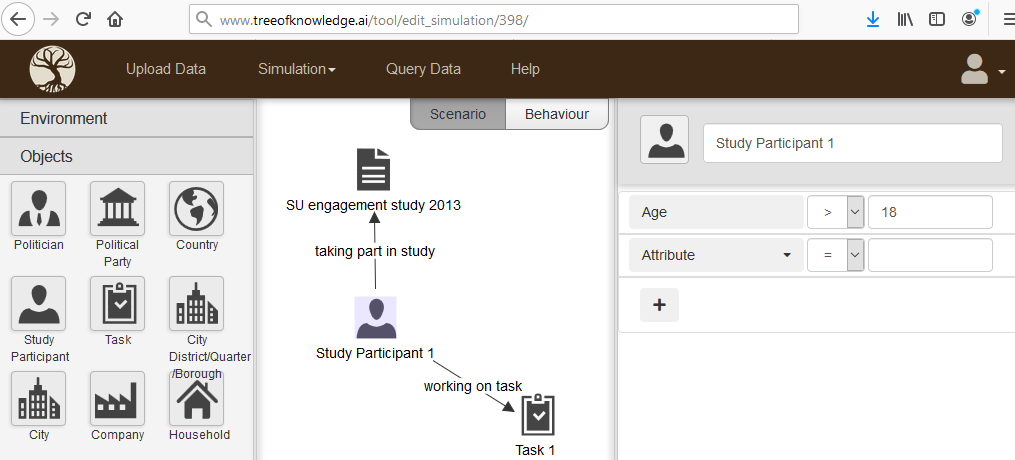}
\caption{\label{fig:fig1} Screenshot from TreeOfKnowledge \newline From the left panel the three agents 'Study', 'Study Participant' and 'Task' were added to the simulation (through drag-and drop). By clicking on 'Study Participant 1' the right panel opens which allows you to make some changes e.g., add filters.}
\label{fig:screenshot_editor}
\end{figure}

We remember that we have some data that could help us fit and evaluate the new behaviour rule. The data is from a study where participants were asked to specify their hunger-level (amongst other things) and then given some task; the participants’ engagement with the task was then also captured in the data. We upload this data to Tree of Knowledge and model the study scenario by drag-and-dropping agents into the simulation, and connecting them with arrows – see Figure ~\ref{fig:screenshot_editor}. Next, we specify that we want to fit and evaluate the new rule and click run. The rule learning happens completely automatically. At the end we are shown a probability distribution for the parameter X and a score which indicates if the model improved with the new rule.

But how could a study on engagement allow us to learn this new happiness rule? The participants’ happiness was not even captured in this study!

One strength of using agent-based models is that a behaviour rule can be learned even if it only plays an indirect role or if there are opposing/self-enforcing mechanisms at play. In our example other users had previously made rules for how someone’s happiness influences their engagement with a task. This indirect connection from hunger to happiness to engagement allows us to nevertheless learn the new rule. For further details on behaviour rules, see Appendix~\ref{app:behaviour_rules}.

\subsection{Likelihood-free Bayesian inference} \label{ssec:bayesian_inference}

As mentioned previously the rule learning consists of two steps: fitting rules and scoring the performance of the fitted rules. The rule fitting learns the unknown parameters of the rules with the formalism of likelihood-free Bayesian inference.

Likelihood-free Bayesian inference is used in various fields such as evolutionary biology \cite{Hickerson:2010, Segelbacher:2001}, epidemiology \cite{Sisson+Fan+Tanaka:2007, Tanaka+Francis+Luciani+Sisson:2006} and systems biology \cite{Toni+Welch+Strelkowa+Ipsen+Strumpf:2009, Toni+Stumpf:2010} for learning model parameters of generative models, such as agent-based models. This framework learns the model parameters by running many simulations - each with different parameter settings - and checking for which parameter settings the simulation was the most accurate \cite{Hartig+Calabrese+Reineking+Wiegand+Huth:2011}. The technical details on how we determine the accuracy of a model are given in Appendix~\ref{app:model_scoring}.

Unlike most optimisation methods, likelihood-free Bayesian inference does not return a single, best-performing parameter value. Instead it returns a probability distribution. If, for example, the system is learning the unknown parameter X - a value we defined to be between -10 and 10, then the result is a probability distribution over that range. This probability distribution contains information about both the best performing (most likely) happiness-decrease value X – the maximum of the distribution – as well as the uncertainty we have for this value – the width of the distribution. 

Due the rigorous Bayesian formalism, the learned probability distributions exactly capture the uncertainty of the parameters due to lack of data or non-conclusive data \cite{Bolstad:2013}. As some parameters might affect each other, the user has to choose which parameters are learned together.

\subsection{Learning from multiple models/datasets} \label{ssec:multiple_models}

We can however do much better! Instead of learning a parameter from one dataset and corresponding model, it can be learned from many!

Usually, a user uploads data to TreeOfKnowledge because they are interested in learning about the real-world system that generated this data. Therefore, often their next step is to model the the system/scenario that generated the data. Over time, users will have uploaded and modelled many different datasets involving one or multiple humans. In each of these models the human agent(s) are in a different scenario e.g. one user uploaded and modelled data from an observational study that observed the purchasing decisions of pedestrians in Tokyo, another user uploaded and modelled data from a study that observed students' retention of information.

We could imagine that in all of these scenarios someone's hunger might have a role on their happiness. TreeOfKnowledge can automatically test this by re-running all of the models, this time however, with the human agent(s) having the additional happiness rule we just created. For each of these models/scenarios we observe if the addition of our new rule improves its performance score or not. These individual scores are then combined to an overall score. The calculation of performance scores is described in Appendix~\ref{app:model_scoring}.

\subsubsection{Robustness}
This ability to test microscopic behaviour in many different scenarios is unique to agent-based behaviour learning and is probably its most powerful feature. By testing a behaviour in increasingly many different scenarios we can become increasingly certain of its general validity. We hope that over time, the TreeOfKnowledge users will figure out the set of agent behaviours that performs best in all scenarios. Having been tested in all these scenarios, we can then claim these results to be very robust. 

\subsubsection{Accuracy}
Before a rule is tested it is first fit to the data i.e. all unknown parameters are learned. In section \ref{ssec:bayesian_inference} we saw how likelihood-free Bayesian inference is used to learn the probability distribution for an unknown parameter. This probability distributions captures the exact knowledge and uncertainty of the model and data on which it was learned. 

We can go one step further and learn the probability distribution on all models. We can for instance learn a probability distribution for the unknown parameter X (the decrease in happiness from being hungry) from each model involving one or multiple humans. Using Equation~\ref{eq:1} we can combine then combine these probability distributions to an overall probability distribution for X. 
\begin{equation}
P(H|E_1, E_2, \dots, E_n) = P(H) * \frac{P(E_1|H)}{P(E_1)}* \frac{P(E_2|H)}{P(E_2)} * \dots * \frac{P(E_n|H)}{P(E_n)}\label{eq:1}
\end{equation}
This overall distribution accurately combines the learnings from the all of the probabity distributions. The knowledge and uncertainty of the overall distribution captures the exact knowledge and uncertainty we get from considering all the models and data. 

\begin{figure}[t!]
\centering
\includegraphics[width=1.0\textwidth]{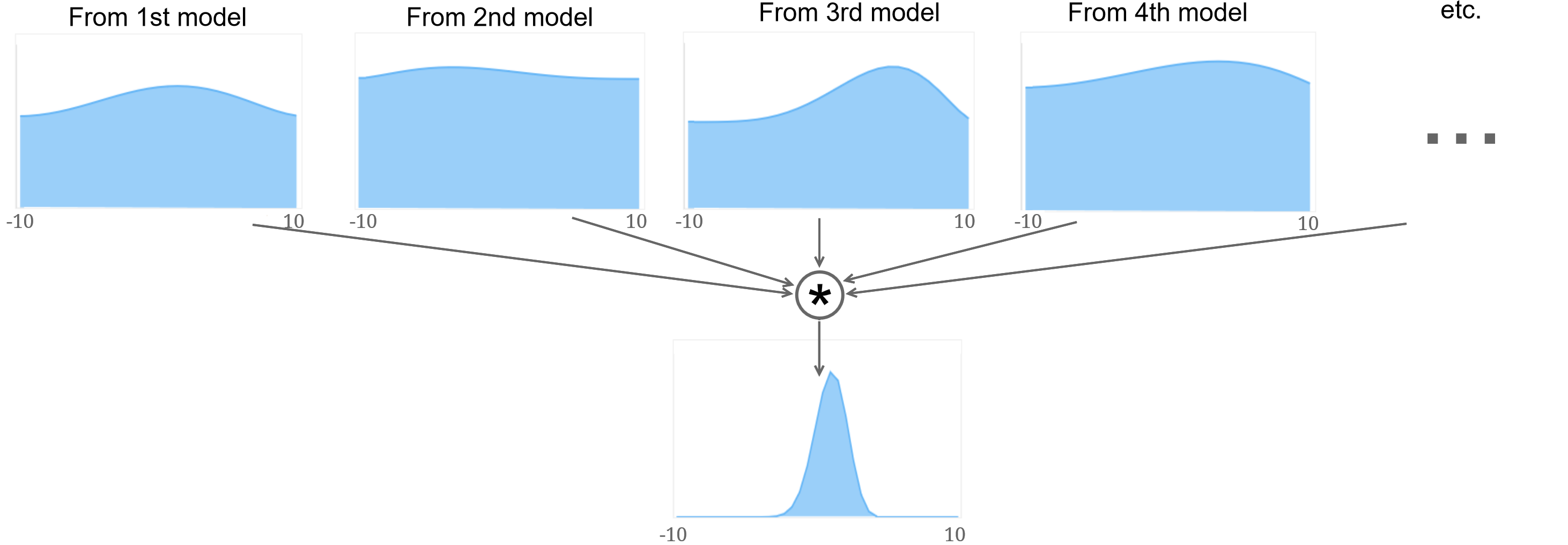}
\caption{\label{fig:fig2} Illustration of the combining of probability distributions.}
\label{fig:multiplying_posteriors}
\end{figure}

The probability distribution of X might be flat (i.e. uninformative) for an individual model and data. E.g. if, for this model, happiness does not have a big effect on the measured outcomes or if the data is not very conclusive. By combining the probability distributions from many models, we however also combine the learnings from these models. The overall distribution is therefore often much more informative and accurate - see Figure~\ref{fig:multiplying_posteriors}.

By exactly capturing the learnings from many different datasets, studies and settings, the overall probability distributions are significantly more robust and accurate than is possible with conventional statistics.

\subsection{Making predictions} \label{ssec:predictions}

The accurate probability distributions can then be used to make accurate predictions.

In TreeOfKnowledge predictions are made by specifying a scenario and running a simulation to see how this scenario unfolds. The probabilistic/Monte-Carlo simulation uses the latest probability distributions for the rule parameters and accurately propagates their uncertainties through the simulation. The simulation results are themselves probability distributions that accurately reflect our uncertainty in the outcome given our present knowledge of the world.

We expect that predictions become more precise as users add and model datasets.

\subsection{World views} \label{ssec:world_views}
In TreeOfKnowledge the full set of all behaviour rules for all agents is called a ‘world view’. If you believe that some behaviour is governed by other rules, you may create a new world view by copying an existing world view and then replacing or adding rules. After the parameters of a world view have been learned, the world view automatically gets a score for how well it matches reality i.e. how well it scores on all models. Users may choose freely which world view they would like to use for modelling their system.

%% -- Discussion ---------------------------------------------------------------

\section{Discussion} \label{sec:discussion}
\subsection{Rigorous framework for reasoning} \label{ssec:framework_for_reasoning}

Agent-based behaviour learning provides a rigorous framework for reasoning about our complex and uncertain world. Predictions are mathematically derived from a world view (i.e. a set of behaviour rules) and the uploaded data. People questioning predictions can either change the world view or upload new data, but must otherwise accept them as true. 

\subsection{Comparison with existing methods} \label{ssec:comparison}

There are various fields that deal with learning the behaviour rules of complex agents such as humans. These fields include: psychology, behavioural economics, ecology, etc. The most used method in these fields is hypothesis testing, however many other statistical methods also get used. 

\subsubsection{Easy to use and interpret}
Statistics and machine learning methods/models generally require knowledge of the respective field to use and interpret them correctly. Many are very hard to interpret even for domain experts, as they learn very high-dimensional patterns/relationships. 
For TreeOfKnowledge, great attention was paid to making it as intuitive and easy to use as possible. Users only have to understand the rules and probability distributions. Also, the models can easily be inspected – an interface allows you to watch how a simulation unfolds.

\subsubsection{Collaborative}
When users in Tree of Knowledge create a new model or add new behaviours to an agent, they are simultaneously also contributing to a big model which grows in scope and accuracy. Collaborative model building is very rare in statistics and machine learning. 

\subsubsection{Learning complex behaviours}
Most of statistics and machine learning can only analyse directly observed behaviour. In psychology therefore much work goes into designing experiments that isolate a single behaviour and avoid confounding factors as much as possible. This is a very difficult thing to do, because in our complex world everything is interacting. 
Agent-based learning on the other hand can learn from any dataset. It learns the parameters of a rule even if it plays a very indirect role (e.g., it changes a not-observed property) or if the behaviour is obscured by opposing or self-enforcing mechanisms. Only probabilistic programming models can do this and they require a lot of programming to set up.

\subsubsection{Rigorous and accurate}
The goal of most statistical analyses is to gain generally valid insights. All methods/models in statistics and machine learning, however, learn from only one single dataset measured in one specific setting. Meta statistics improves this situation by allowing insights from similar studies to be combined. The problem of limited settings however remains and meta statistics is not mathematically rigorous as it requires hand-picked weights.
Agent-based behaviour learning on the other hand provides a rigorous framework for combining learnings from any number of heterogeneous datasets observed in any setting. 
It is very powerful to test a rule in many different scenarios and validate it using many different datasets!

\subsection{Outlook} \label{ssec:outlook}
We believe that agent-based behaviour learning allows social sciences to get a better understanding of the complex systems we live in. We are convinced the insights and predictions made with it are significantly more rigorous and accurate than what has been possible before. We would like to invite any researcher to try the alpha version at \url{www.treeofknowledge.ai}. It shows the framework, but still needs data and models.

%% -- Computational Details -------------------------------------

%% -- Bibliography -------------------------------------------------------------
\bibliographystyle{unsrt}  
\bibliography{tree_of_knowledge}

\begin{thebibliography}{10}

\bibitem{Mitleton-Kelly:2003}
Eve Mitleton-Kelly.
\newblock Ten principles of complexity \& enabling infrastructures.
\newblock {\em Complex Systems and Evolutionary Perspectives on Organisations:
  The Application of Complexity Theory to Organisations}, pages 20--51, 2003.

\bibitem{Arthur:2018}
Brian~W. Arthur.
\newblock {\em The Economy as an Evolving Complex System II}.
\newblock CRC Press, Boca Raton, FL, 2018.

\bibitem{Tesfatsion:2002}
Leigh Tesfatsion.
\newblock Agent-based computational economics: Growing economies from the
  bottom up.
\newblock {\em SSRN Electronic Journal}, 2002.

\bibitem{Arthur:2005}
Brian~W. Arthur.
\newblock Out-of-equilibrium economics and agent-based modelling.
\newblock {\em Handbook of Computational Economics}, 2:1551--1564, 2005.

\bibitem{Heckbert+Baynes+Reeson:2010}
Scott Heckbert, Tim Baynes, and Andrew Reeson.
\newblock Agent-based modeling in ecological economics.
\newblock {\em Annals of the New York Academy of Sciences}, 1185:39--53, 2010.

\bibitem{Graebner:2015}
Claudius Gräbner.
\newblock Methodology does matter: About implicit assumptions in applied formal
  modelling. the case of dynamic stochastic general equilibrium models vs
  agent-based models.
\newblock {\em MPRA Paper}, 63003, 2015.

\bibitem{Colander+Holt+Rosse:2004}
David Colander, Richard P.~F. Holt, and J.~Barkley Rosser.
\newblock The changing face of mainstream economics.
\newblock {\em Review of Political Economy}, 16(4):485--499, 2004.

\bibitem{Deissenberg+Van_der_Hoog+Dawid:2008}
Christophe Deissenberg, Sander~Van der Hoog, and Herbert Dawid.
\newblock Eurace: A massively parallel agent-based model of the european
  economy.
\newblock {\em Applied Mathematics and Computation}, 204(2):541--552, 2008.

\bibitem{Macy+Willer:2002}
Michael~W. Macy and Robert Willer.
\newblock From factors to actors: Computational sociology and agent-based
  modeling.
\newblock {\em Annual Review of Sociology}, 28(1):143--166, 2002.

\bibitem{Smith+Conrey:2007}
Eliot~R. Smith and Frederica~R. Conrey.
\newblock Agent-based modeling: A new approach for theory building in social
  psychology.
\newblock {\em Personality and Social Psychology Review}, 11(1):87--104, 2007.

\bibitem{Premo:2006}
L.~S. Premo.
\newblock Agent-based models as behavioral laboratories for evolutionary
  anthropological research.
\newblock {\em Arizona Anthropologist}, 17:91–--113, 2006.

\bibitem{Heppenstall+Crooks+See+Batty:2011}
Alison~J. Heppenstall, Andrew~T. Crooks, Linda~M. See, and Michael Batty.
\newblock {\em Agent-Based Models of Geographical Systems}.
\newblock Springer Science \& Business Media, Berlin, 2011.

\bibitem{Grimm+Railsback:2005}
Volker Grimm and Steven~F. Railsback.
\newblock {\em Individual-Based Modeling and Ecology}.
\newblock Princeton University Press, Princeton, NJ, 2005.

\bibitem{Coakley+Smallwood+Holcombe:2006}
Simon Coakley, Rod Smallwood, and Mike Holcombe.
\newblock From molecules to insect communities - how formal agent based
  computational modelling is uncovering new biological facts.
\newblock {\em Scientiae Mathematicae Japonicae Online}, e-2006:765--–778,
  2006.

\bibitem{Fullstone+Wood+Holcombe+Battaglia:2015}
Gavin Fullstone, Jonathan Wood, Mike Holcombe, and Giuseppe Battaglia.
\newblock Modelling the transport of nanoparticles under blood flow using an
  agent-based approach.
\newblock {\em Scientific Reports}, 5(1), 2015.

\bibitem{Shen+Wang+Hao:2006}
Weiming Shen, Lihui Wang, and Qi~Hao.
\newblock Agent-based distributed manufacturing process planning and
  scheduling: A state-of-the-art survey.
\newblock {\em IEEE Transactions on Systems, Man and Cybernetics},
  36(4):563--577, 2006.

\bibitem{Farmer+Foley:2009}
J.~Doyne Farmer and Duncan Foley.
\newblock The economy needs agent-based modelling.
\newblock {\em Nature}, 460(7256):685–--86, 2009.

\bibitem{Janssen+Ostrom:2006}
Marco~A. Janssen and Elinor Ostrom.
\newblock Empirically based, agent-based models.
\newblock {\em Ecology and Society}, 11(2), 2006.

\bibitem{Srbljinovic+Skunca:2003}
Armano Srbljinović and Ognjen Škunca.
\newblock An introduction to agent-based modelling and simulations of social
  processes.
\newblock {\em Interdisciplinary Description of Complex Systems}, 1(1-2):1--8,
  2003.

\bibitem{Carley:1996}
Kathleen~M. Carley.
\newblock Validating computational models.
\newblock {\em CASOS working paper, Carnegie Mellon University}, 1996.

\bibitem{Wolf:1986}
Frederic~M. Wolf.
\newblock {\em Meta-analysis: Quantitative Methods for Research Synthesis}.
\newblock Sage Publications, Thousand Oaks, CA, 1986.

\bibitem{Schulze:2004}
Ralf Schulze.
\newblock {\em Meta-Analysis - A Comparison of Approaches}.
\newblock Hogrefe Publishing, Göttingen, 2004.

\bibitem{Breakwell+Smith+Wright:2012}
Glynis~M. Breakwell, Jonathan~A. Smith, and Daniel~B. Wright.
\newblock {\em Research Methods in Psychology}.
\newblock Sage Publications, Thousand Oaks, CA, 4th edition, 2012.

\bibitem{Senn:2012}
Stephen Senn.
\newblock Seven myths of randomisation in clinical trials.
\newblock {\em Statistics in Medicine}, 32(9):1439--1450, 2012.

\bibitem{Hickerson:2010}
M.~J. Hickerson, B.~C. Carstens, J.~Cavender-Bares, K.~A. Crandall, C.~H.
  Graham, J.~B. Johnson, L.~Rissler, P.~F. Victoriano, and A.~D. Yoder.
\newblock Phylogeography’s past, present, and future: 10 years after avise,
  2000.
\newblock {\em Molecular Phylogenetics and Evolution}, 54(1):291–--301, 2010.

\bibitem{Segelbacher:2001}
Gernot Segelbacher, Samuel~A. Cushman, Bryan~K. Epperson, Marie-Josée Fortin,
  Olivier Francois, Olivier~J. Hardy, Rolf Holderegger, Pierre Taberlet,
  Lisette~P. Waits, and Stéphanie Manel.
\newblock Applications of landscape genetics in conservation biology: Concepts
  and challenges.
\newblock {\em Conservation Genetetics}, 11(2):375–--385, 2001.

\bibitem{Sisson+Fan+Tanaka:2007}
S.~A. Sisson, Y.~Fan, and Mark~M. Tanaka.
\newblock Sequential monte carlo without likelihoods.
\newblock {\em Proceedings of the National Academy of Sciences of the U. S.
  A.}, 104(6):1760–--1765, 2007.

\bibitem{Tanaka+Francis+Luciani+Sisson:2006}
Mark~M. Tanaka, Andrew~R. Francis, Fabio Luciani, and S.~A. Sisson.
\newblock Estimating tuberculosis transmission parameters from genotype data
  using approximate bayesian computation.
\newblock {\em Genetics}, 173:1511–--1520, 2006.

\bibitem{Toni+Welch+Strelkowa+Ipsen+Strumpf:2009}
Tina Toni, David Welch, Natalja Strelkowa, Andreas Ipsen, and Michael~P.H.
  Stumpf.
\newblock Approximate bayesian computation scheme for parameter inference and
  model selection in dynamical systems.
\newblock {\em J. R. Soc. Interface}, 6(31):187–--202, 2009.

\bibitem{Toni+Stumpf:2010}
Tina Toni and Michael~P.H. Stumpf.
\newblock Simulation-based model selection for dynamical systems in systems and
  population biology.
\newblock {\em Bioinformatics}, 26(1):104--–110, 2010.

\bibitem{Hartig+Calabrese+Reineking+Wiegand+Huth:2011}
Florian Hartig, Justin~M. Calabrese, Björn Reineking, Thorsten Wiegand, and
  Andreas Huth.
\newblock Statistical inference for stochastic simulation models - theory and
  application.
\newblock {\em Ecology Letters}, 14(8):816--827, 2011.

\bibitem{Bolstad:2013}
William~M. Bolstad.
\newblock {\em Introduction to Bayesian Statistics}.
\newblock John Wiley \& Sons, Hoboken, NJ, 2nd edition, 2013.

\bibitem{wikipedia}
Comparison of agent-based modeling software.
\newblock
  \url{https://en.wikipedia.org/wiki/Comparison_of_agent-based_modeling_software}.
\newblock Accessed: 2020-05-01.

\end{thebibliography}

%% -- Appendix  --------------------------------------------------------

\newpage

\begin{appendix}

\section{Agent types and agents} \label{app:agent_types}

\begin{figure}[t!]
\centering
\includegraphics[width=0.3\textwidth]{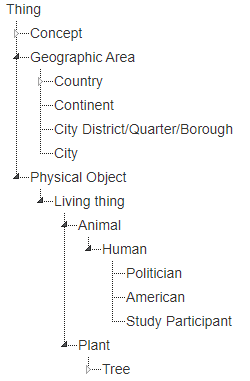}
\caption{\label{fig:fig3} Formal Ontology of agent types\\
Sorting agents into such an ontological tree allows the behaviour of the more general agent types to be learned from a wide variety of datasets. This tree will grow as users add new agent types to TreeOfKnowledge.}
\label{fig:agent_type_hierachy}
\end{figure}

TreeOfKnowledge has a central repository of agent types, which users use to model their systems of interest. These agent types are sorted into a formal ontology/taxonomic tree – see Figure ~\ref{fig:agent_type_hierachy}. If an agent type is missing from the central repository, users may add it. In theory you could make an agent type for any object/concept that you could use as the noun of a sentence. 

An agent (e.g., Alice) is an instantiation/grounding of an agent type (e.g., Politician). Every agent has attributes (e.g., age, height) and relations to other agents (e.g ‘is born in country X’, ‘is married to Y’). The list of attributes and relations an agent type is inherited from its parents in the formal ontology - see Figure~\ref{fig:agent_type_hierachy}. Where required, users can also create additional attributes or relations.

During a simulation the attributes and relations of the simulated agents change over time. How they change is specified by a set of behaviour rules. The set of behaviour rules of an agent is also inherited from its parents in the formal ontology. This allows general rules (e.g., those used to simulate the metabolism of an animal) to be learned from a wide range of datasets and more specific rules (e.g., relating to a politician changing party affiliation) to be only learned on the relevant datasets.

\section{Behaviour Rules} \label{app:behaviour_rules}

TreeOfKnowledge uses behaviour rules to specify agent behaviour. Many other agent-based modelling tools use programmed code \cite{wikipedia}. We believe that rules have the advantages of being more modular/extensible and of being easier to understand by a wide audience.

The rules are manually created by users. The system, however, learns which of these rules is indeed correct and learns all unknown parameters and rule probabilities. It would be possible to have entirely machine-learned rules, but man-made rules are significantly more legible, usually have gone through a basic sense check and do not overfit to the data.

Any type of rule or mathematical expression can be expressed as behaviour rule, including logic-expressions, differential equations, calculations, string transformations, etc. Additionally, behaviour rules can be:
\begin{itemize}
  \item Probabilistic or certain.
  \item Have unknown parameters or not.
  \item Have an if-condition or not.
\end{itemize}

\begin{figure}[t!]
\centering
\includegraphics[width=0.95\textwidth]{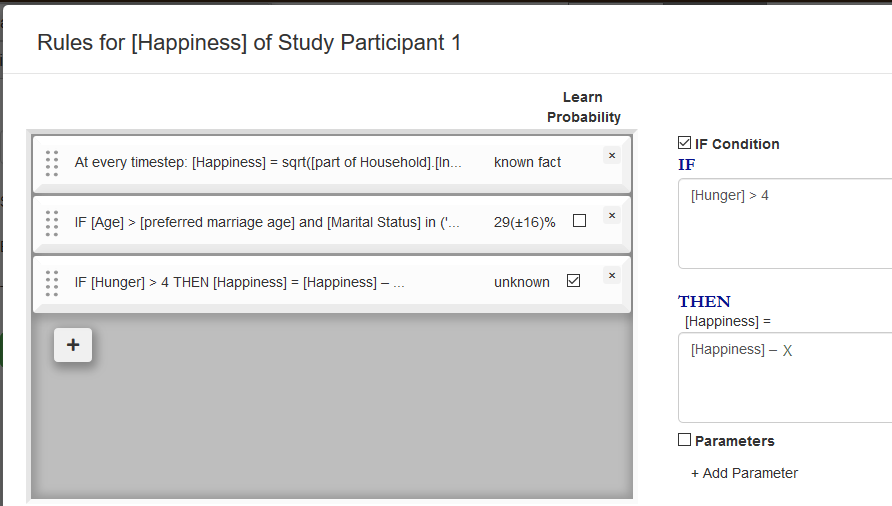}
\caption{\label{fig:fig4} Screenshot from the rule-inspection window in TreeOfKnowledge\\
The left side shows the list of the rules affecting a Study Participant’s happiness. The tick for the third rule indicates that it’s probability will be learned. If you click on one of these rules, the details of the rule are displayed on the right (here the third rule was clicked).}
\label{fig:rule_inspection}
\end{figure}

Some examples:\\
In section \ref{ssec:model_building} of this paper the following rule is mentioned: “if someone’s hunger-level exceeds 4, then their happiness is reduced by X”. Written in TreeOfKnowledge’s notation this is “IF [Hunger] > 4 THEN [Happiness] = [Happiness] - X" – see Figure  ~\ref{fig:rule_inspection}. This is a certain rule (i.e., non-probabilistic), but has the unknown parameter X, which will be learned by the system.

Another rule affecting someone’s happiness might be "IF [lives in household].[Is in dept] == True THEN [Happiness] = [Happiness] - Y". This reads as “if the household someone lives in is in dept, then their happiness will be reduced by Y”. As we are not at all sure if this rule is correct, we decide to make it a probabilistic rule i.e., if its condition is fulfilled it will still only be executed with a certain probability. For this rule the system now has two parameters to learn: Y and the rule probability. 

Also, the above rule contains an example of the use of an attribute from a related agent. [lives in household].[Is in dept] refers to the attribute [Is in dept] of the household that is connected to the person with the relation [lives in household]. In general, an agent’s behaviour rules can use the attributes of any agent that are up to three relations away.

Other rules might be physics formulas or known transformations such as "[Height (meters)] = 3.281 * [Height (feet)]”. These formulas usually are certain rules (non-probabilistic), don’t have any if-condition and also don’t have any unknown parameters. We know of no type of specific knowledge that could not be entered into TreeOfKnowledge either as rule or as datapoint.

\section{Model Scoring} \label{app:model_scoring}

Agent-based behaviour learning depends on users uploading datasets to TreeOfKnowledge and modelling the scenario/setting in which the dataset was observed. In this section we describe how a model’s performance is evaluated by comparing simulation results to the real-world data. 

For any model, the system can automatically calculate the model's performance score by comparing the simulated values to the real observations from the real world. Section \ref{ssec:bayesian_inference} describes how this score is used both for likelihood-free Bayesian inference of parameters and for determining if a rule improves a model’s performance.

To see how TreeOfKnowledge calculates this score, we will look at a simple model that contains only one agent – Sri Lanka. This model simulates Sri Lanka for the years 2010–2017 using timesteps of one year. As we can see in Figure ~\ref{fig:model_scoring_diagram}, the simulation scoring is done in 4 steps:

\begin{figure}[t!]
\centering
\includegraphics[width=1.0\textwidth]{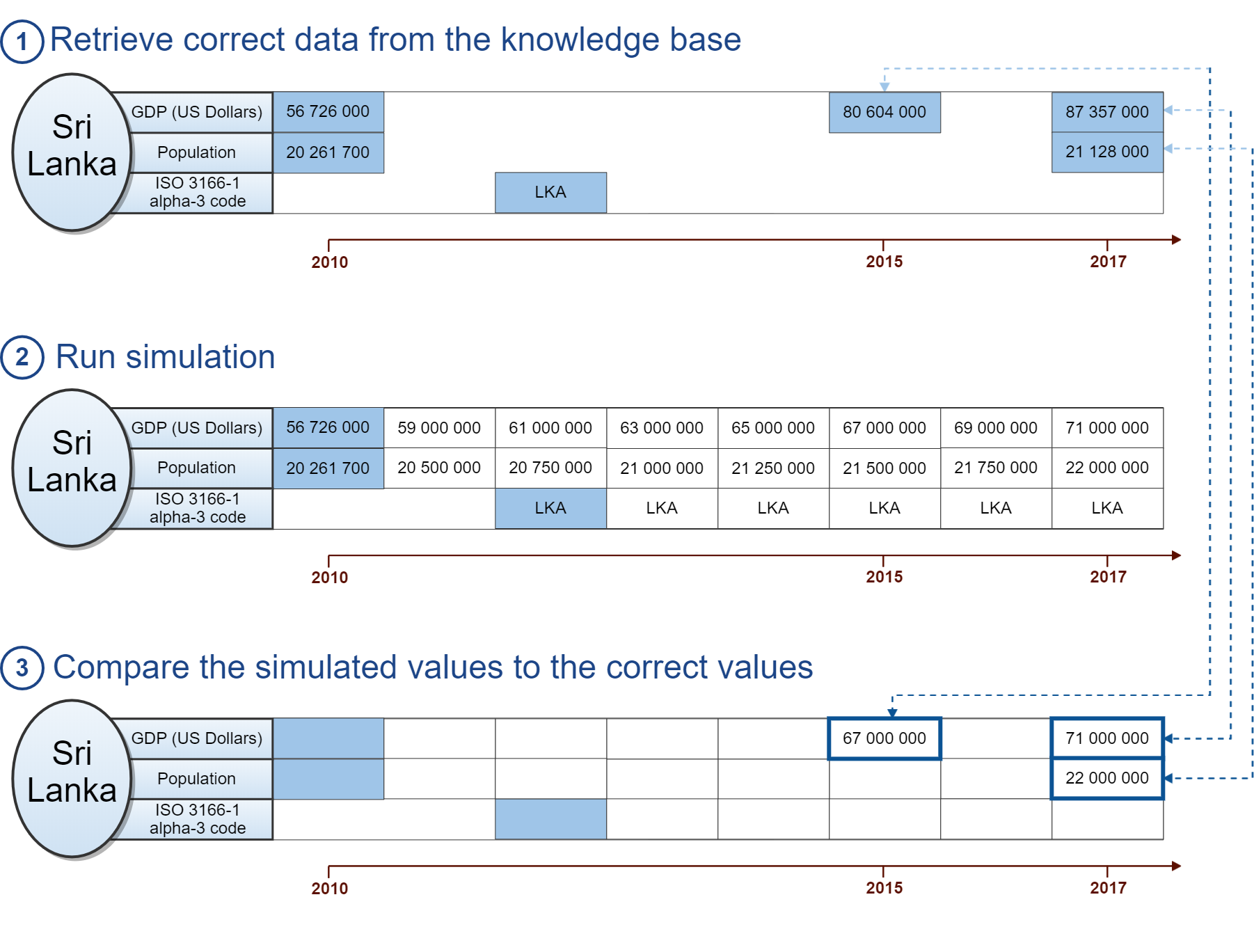}
\caption{\label{fig:fig5} This diagram illustrates how the model scoring works for a model with the single agent Sri Lanka.}
\label{fig:model_scoring_diagram}
\end{figure}

\begin{enumerate}
    \item Retrieve correct data from the knowledge base \newline
As we uploaded data from the real system (here: facts about Sri Lanka) to TreeOfKnowledge, this data was automatically integrated with previously uploaded facts about Sri Lanka in TreeOfKnowledge’s knowledge base. In the first step, all relevant facts about Sri Lanka are retrieved from the knowledge base and combined to a timeline.

\item Run simulation\newline
In the second step, the first values from each of Sri Lanka’s attributes (here: GDP, Population and ISO code) is used to initialize a simulation. Then the simulation is run by executing the behaviour rules for every timestep.

\item Compare the simulated values to correct values\newline
Where possible, the simulated values are compared to the correct values from the knowledge base. The score for this simulation is the average score from the individual value-comparisons.

\item Repeat\newline
The simulations are usually probabilistic, which means that they unfold differently every time. To account for this, the simulation is run many times to calculate an average score.

Also, if a model involves a more general agent type, say ‘country’ instead of ‘Sri Lanka’, then the model will be evaluated at least once for every country found in the knowledge base.
\end{enumerate}

\end{appendix}

%% -----------------------------------------------------------------------------

\end{document}